\documentclass[conference]{IEEEtran}

\usepackage{cite}
\usepackage[pdftex]{graphicx}
\usepackage[cmex10]{amsmath}
\usepackage[caption=false,font=footnotesize]{subfig}

\usepackage{amssymb}
\usepackage{amsthm}
\usepackage{color}
\usepackage{bm}
\usepackage{url}
\usepackage{booktabs}

\newtheorem{thm}{Theorem}

\newtheorem{prop}[thm]{Proposition}

\newcommand{\mbf}{\mathbf}

\newcommand{\bs}{\boldsymbol}

\begin{document}
%
\title{Recovery of Piecewise Smooth Images\\ from Few Fourier Samples}

\author{\IEEEauthorblockN{Greg Ongie}
\IEEEauthorblockA{Department of Mathematics\\
University of Iowa\\
Iowa City, Iowa 52242\\
Email: gregory-ongie@uiowa.edu}
\and
\IEEEauthorblockN{Mathews Jacob}
\IEEEauthorblockA{Department of Electric and Computer Engineering\\
University of Iowa\\
Iowa City, Iowa 52242\\
Email: mathews-jacob@uiowa.edu}}


%


\maketitle

\begin{abstract}
We introduce a Prony-like method to recover a continuous domain 2-D piecewise smooth image from few of its Fourier samples. Assuming the discontinuity set of the image is localized to the zero level-set of a trigonometric polynomial, we show the Fourier transform coefficients of partial derivatives of the signal satisfy an annihilation relation. We present necessary and sufficient conditions for unique recovery of piecewise constant images using the above annihilation relation. We pose the recovery of the Fourier coefficients of the signal from the measurements as a convex matrix completion algorithm, which relies on the lifting of the Fourier data to a structured low-rank matrix; this approach jointly estimates the signal and the annihilating filter. Finally, we demonstrate our algorithm on the recovery of MRI phantoms from few low-resolution Fourier samples.  
\end{abstract}


\section{Introduction}
The recovery of continuous domain parametric representations from few measurements using harmonic retrieval/linear prediction has received considerable attention in signal processing since Prony's seminal work\cite{stoica1997introduction,cheng2003review}. Extensive research has been devoted to the recovery of finite linear combination of exponentials with unknown continuous frequencies as well as linear combination of Diracs and non-uniform 1-D splines with unknown locations/knots \cite{vetterli2002sampling,maravic2005sampling,dragotti2007sampling}. Recently, convex algorithms that minimize atomic norm were also introduced to recover such continuous signals \cite{bhaskar2011atomic,candes2013super}; these methods are off-the-grid continuous generalizations of compressed sensing theory, which can avoid discretization errors and hence are potentially more powerful. However, the direct extension of the above Prony-like and convex off-the-grid methods to 2-D piecewise smooth images is not straightforward. Specifically, the partial derivatives of piecewise smooth images can be thought of as a linear combination of a continuum of Diracs supported on the curves separating the regions, which current methods are not designed to handle. 


Recently, Pan \emph{et al.} \cite{pan2013sampling} introduced a complex analytic signal model for continuous domain 2-D images, such that the complex derivatives of the image are supported on a curve. Under the assumption that the curve is the zero-set of a band-limited function, the authors show the Fourier transform of the complex derivative of such a signal is annihilated by convolution with the Fourier coefficients of the band-limited function. This property is used to extend the finite-rate-of-innovation model \cite{vetterli2002sampling} to this class of 2-D signals. This work has several limitations. One problem is that a complex analytic signal model is not realistic for natural images (e.g., the only real-valued analytic functions are constant functions). In addition, if we are only measuring a finite number of Fourier samples, one can choose analytic functions such that the all of these coefficients vanish; i.e. the recovery of the signal from few Fourier coefficients of an arbitrary signal in this class is ill-posed. 

In this work, we address the above limitations by proposing an alternative signal model based on the a more realistic class of piecewise smooth functions. We show that we can generalize the annihilation property in \cite{pan2013sampling} to this class of functions, such that it is possible to recover an exact continuous domain representation of the edge set from few Fourier samples. We also determine necessary and sufficient conditions on the number of Fouier samples required for perfect recovery of the edge set in the caseof  piecewise constant images.

To recover the full signal, we propose a single-step convex algorithm which can be thought of as jointly estimating the edge set and image amplitudes. This is fundamentally different than the two stage approach proposed in \cite{pan2013sampling}, which we also investigated for super-resolution MRI in \cite{isbi2015}. Motivated by recently proposed algorithms for calibration-free parallel MRI recovery \cite{sake,loraks}, our new approach is based on the observation that the ideal Fourier samples of the signal lift to a structured low-rank matrix, which allows us to pose the recovery as a low-rank structured matrix completion problem. We demonstrate our algorithm on the recovery of MR phantoms from low-resolution Fourier samples.

\section{Signal Model}

\subsection{Piecewise smooth signals}

In this paper, we consider the general class of 2-D \emph{piecewise smooth} functions:
\begin{equation}
\label{maineqn}
f(\mbf r) = \sum_{i=1}^n g_i(\mbf r)~ \chi_{\Omega_i}(\mbf r),~~\forall \mbf r = (x,y)\in[0,1]^2,
\end{equation}
where  $\chi_{\Omega}$ denotes the characteristic function of the set $\Omega$:
\begin{equation}
\chi_\Omega(\mathbf r) = 
\begin{cases}
1 & \text{if } \mbf r = (x,y) \in \Omega\\
0 & \text{else}.
\end{cases}
\label{eq:char_func}
\end{equation} 
Here we assume each $\Omega_i\subset [0,1]^2$ is a simply connected region with piecewise smooth boundary $\partial \Omega_i$. The functions $g_i$ in (\ref{maineqn}) are smooth functions that vanish under of a collection of constant coefficient differential operators $\mathbf D = \{D_1,...,D_N\}$ within the region $\Omega_i$:
\begin{equation}
D_j~ g_i(\mbf r) = 0, \forall \mbf r \in \Omega_i; ~j=1,..,N.
\end{equation}
We now show that the above class of functions is fairly general and includes many well-understood image models by appropriately choosing the set of differential operators $\mathbf D$.

\subsubsection{\underline{Piecewise constant images}} We set $\mbf D$ to 
\[
\mbf D = \nabla = \{\partial_x, \partial_y\}.
\] 
Note that $\mbf D g = 0$ if and only if $g = c_i$ for $c_i\in\mathbb{C}$. Hence, (\ref{maineqn}) reduces to the well-known piecewise constant image model:
\begin{equation}
f(\mbf r) = \sum_{i=1}^n c_i\,\chi_{\Omega_i}(\mbf r),~~\forall \mbf r = (x,y)\in[0,1]^2,
\end{equation}
This case will be the primary focus of this work due to its simplicity and provable guarantees.

\subsubsection{\underline{Piecewise analytic images}} Choosing $\mbf D = \partial_{\bar{z}} = \partial_x + j \partial_y$, then $\mbf D g = 0$ if and only if $g$ is complex analytic. Hence this model is equivalent to the one proposed in \cite{pan2013sampling}. As described above, this signal model is not very realistic for natural images. 

\subsubsection{\underline{Piecewise harmonic}} Both the above cases consider only first-order differenial operators. One choice of a second-order differential operator is the Laplacian $\mbf D = \Delta = \partial^2_{xx} + \partial^2_{yy}$. Then $\mbf D g = 0$ if and only if $g$ is harmonic. 


\subsubsection{\underline{Piecewise linear images}} If we consider all second order partial derivatives $\mbf D = \{\partial^2_{xx}, \partial^2_{xy}, \partial^2_{yy}\}$, then $\mbf D g = 0$ if and only if $g$ is linear, i.e. $g(\mbf r) = \langle \mbf a,\mbf r\rangle + b$, for $\mbf a \in \mathbb{C}^2$, $b\in \mathbb{C}$, and so $f$ has the expression
\begin{equation}
f(\mbf r) = \sum_{i=1}^n \left(\langle\mbf a_i,\mbf r\rangle + b_i\right)~ \chi_{\Omega_i}(\mbf r),~~\forall \mbf r = (x,y)\in[0,1]^2.
\end{equation}

\subsubsection{\underline{Piecewise polynomial}} Generalizing the above case we may consider all $n$th order partial derivatives $\mbf D = \{\partial^\alpha\}_{|\alpha|=n}$ where $\alpha$ is a multi-index. then $\mbf D g = 0$ if and only if $g$ is a polynomial of degree at most $(n-1)$.

We will show that under certain assumptions on the edge set $C=\cup_{i=1}^n\partial \Omega_i$, the Fourier transform of derivatives of a piecewise smooth signal specified by (\ref{maineqn}) satisfies an annihilation property. This will enable us to recover an exact continuous domain representation the edge set $C$ of a piecewise smooth signal from finitely many of its Fourier samples by solving a linear system.

\subsection{Trigonometric polynomials and curves}
Following \cite{pan2013sampling}, we will assume the edge set $C$ to be the zero-set of a band-limited periodic trigonometric polynomial
\begin{equation}
\mu(\mbf r) = \sum_{\mbf k\in\Lambda} c[\mbf k]\, e^{j2\pi\langle \mbf k,\mbf r \rangle},\quad \forall \mbf r\in{[0,1]}^2,
\label{eq:trigpoly}
\end{equation}
where $c[\mbf k]\in\mathbb{C}$ and $\Lambda$ is any finite subset of $\mathbb{Z}^2$; we call any function $\mu$ described by \eqref{eq:trigpoly} a \emph{trigonometric polynomial}, and the zero-set $C: \{\mu=0\}$ a \emph{trigonometric curve}. We also define the \emph{degree} of a trigonometric polynomial $\mu$ to be the dimensions of the smallest rectangle that contains the frequency support set $\Lambda$, denoted as $deg(\mu) = (K,L)$. For trigonometric polynomials $\mu$ and $\nu$, we say $\nu$ \emph{divides} $\mu$ or $\nu~|~\mu$ if $\mu = \nu \cdot \gamma$ where $\gamma$ is another trigonometric polynomial.

Using elementary results from algebraic geometry, we may show there is a unique minimal degree trigonometric polynomial associated with any trigonometric curve $C$, which we call the \emph{minimal polynomial} for $C$: 

\begin{prop}
For every trigonometric curve $C$ there is a unique (up to scaling) trigonometric polynomial $\mu_0$ with $C:\{\mu_0 = 0\}$ such that for any other trigonometric polynomial $\mu$ with $C:\{\mu = 0\}$ we have $deg(\mu_0) \leq deg (\mu)$ and $\mu_0~|~\mu$. 
\end{prop}

The following property of minimal polynomials is also important for our uniqueness results:

\begin{prop}
\label{prop:musquared}
Let $C$ be the zero set of a trigonometric polynomial with minimal polynomial $\mu_0$. Suppose $\nu$ is a trigonometric polynomial such that $\nu = 0$ and $\nabla \nu = 0$ on $C$, then $\mu_0^2~|~\nu$. In particular, $\nabla \mu_0 = 0$ for at most finitely many points on $C$.
\end{prop}

\section{Annihilation property}
We now show that the Fourier transform of the partial derivatives of piecewise smooth signals \eqref{maineqn} satisfy an annihilation property.
\subsubsection{\underline{First order partial derivative operators}}
First we consider the case of a single characteristic function $\chi_\Omega$. Note that since $\chi_\Omega$ is non-smooth at the boundary, its derivatives are only defined in a distributional sense. Letting $\varphi$ denote any test function we have:
\begin{equation}
\langle \partial_x \chi_\Omega, \varphi\rangle  = - \langle \chi_\Omega, \partial_x \varphi\rangle =  -\int_\Omega \partial_x \varphi \, d\mbf r = -\oint_{\partial\Omega} \varphi\, dy
\end{equation}
where the last step follows by Green's theorem. Likewise,
\begin{equation}
\langle \partial_y \chi_\Omega, \varphi\rangle  = \oint_{\partial\Omega} \varphi\, dx
\end{equation}
Hence $\partial_x \chi_\Omega$ and $\partial_y \chi_\Omega$ can be interpreted as a continuous stream of weighted Diracs supported on $\partial\Omega$. In particular, if $\psi$ is any smooth function that vanishes on $\partial\Omega$ then
\begin{equation}
\psi\cdot\partial_x \chi_\Omega = \psi\cdot\partial_y \chi_\Omega = 0
\label{eq:spacedom}
\end{equation}
where equality holds in the distributional sense. Assuming $\psi = \mu$ is a trigonometric polynomial, taking Fourier transforms of \eqref{eq:spacedom} yields the following annihilation relation:

\begin{prop}
Let $f = \chi_\Omega$ with boundary $\partial\Omega$ given by the trigonometric curve $C: \{\mu = 0\}$. Let $D$ be any first order differential operator. Then the Fourier transform of $Df$ is annihilated by convolution with the Fourier coefficients $c[\mbf k],\mbf k \in \Lambda$ of $\mu$, that is
\begin{equation}
  \sum_{\mbf k \in \Lambda} c[\mbf k]\, \widehat{Df}(\bs\omega - 2\pi\mbf k) = 0, ~~\text{for all } \bs \omega \in \mathbb{R}^2.
  \label{eq:annihilation}  
\end{equation}
\end{prop}
Due to the above property, we call $\mu$ an \emph{annihilating polynomial} for $Df$.

It is straightforward to extend the above proposition to piecewise constant functions $f = \sum_{i=1}^n c_i \chi_{\Omega_i}$, provided $\mu = 0$ on the union of the boundaries $C = \cup_{i=1}^n \partial \Omega_i$. Likewise, if $f = g \cdot \partial_\Omega$ where $Dg = 0$, then by the product rule
\[
D f = D g \cdot \chi_\Omega + g \cdot D\chi_\Omega = g \cdot D\chi_\Omega
\]
which has support on $\partial\Omega$ and so, $\mu\cdot D f = 0$, which implies \eqref{eq:annihilation} holds for $f$, and similarly for the linear combination $f = \sum_{i=1}^n g_i \cdot \chi_{\Omega_i}$, where $Dg_i = 0$ for all $i=1,...,n$.


\subsubsection{\underline{Second order partial derivative operators}}

Now consider the case where $D$ is any second-order differential operator. Let $f = g\cdot \chi_\Omega$ where $D g = 0$. We now show that $\mu^2$ is an annihilating polynomial for  $Df$, where $\mu$ is any trigonometric polynomial that annihilates the partial derivatives of $\chi_\Omega$. 

Let $\partial^2 = \partial_2\partial_1$ where $\partial_i \in \{\partial_x, \partial_y\}, i=1,2$. By the product rule we have:
\[
  \partial^2 f = \partial^2 g \cdot \chi_\Omega + \partial_1 g \cdot \partial_2 \chi_\Omega + \partial_2 g \cdot \partial_1 \chi_\Omega + g \cdot \partial^2 \chi_\Omega.
\]
Since $\partial_1 \chi_\Omega$ and $\partial_2 \chi_\Omega$ are annihilated by $\mu$, we have
\[
\mu^2 \cdot \partial^2 f = \chi_\Omega \cdot \mu^2 \cdot \partial^2 g  + g \cdot \mu^2 \cdot \partial^2 \chi_\Omega.
\]
Again by the product rule
\[
  \mu^2 \cdot \partial^2 \chi_\Omega = \partial_2 (\mu^2 \cdot \partial_1 \chi_\Omega) - 2\,\mu \cdot \partial_2 \mu \cdot \partial_1 \chi_\Omega  = 0,
\]
which implies
\[
\mu^2 \cdot \partial^2 f = \chi_\Omega \cdot \mu^2 \cdot \partial^2 g
\]
and so by linearity
\begin{equation*}
\mu^2 \cdot D f = \chi_\Omega \cdot \mu^2 \cdot D g = 0.
\end{equation*}
The above shows that $\mu^2$ is always sufficient to annihilate $D f$, where $D$ is second-order. However, using Prop.\ \ref{prop:musquared}, we may show that when $g$ does not vanish on $\partial\Omega$, then $\mu^2$ is also necessary for annhilation of $Df$, in the sense that if $\nu$ is any other trig polynomial satisfying $\nu \cdot Df = 0$, then $\mu_0^2~|~\nu$, where $\mu_0$ is the minimal polynomial for $\partial\Omega$. If $deg(\mu_0) = (K,L)$, this implies any annihilating polynomial $\nu$ for $Df$ has $deg(\nu) \geq (2K-1,2L-1)$.

\subsubsection{\underline{Partial derivative operators of arbitrary order}}

A similar argument shows that when $D$ is any $n$th order differential operator, and $f = g\cdot \chi_\Omega$ where $D g = 0$, then
\begin{equation}
\mu^n \cdot D f = 0.
\end{equation}
This yields the following annihilation relation for higher-order differential operators:
\begin{prop}
Let $D$ be any $n$th order differential operator. Let $f = g \cdot \chi_\Omega$ with $Dg = 0$ and $\partial\Omega \subset \{\mu = 0\}$ for some trigonometric polynomial $\mu$. Then the Fourier transform of $Df$ is annihilated by convolution with the Fourier coefficients $d[\mbf k], \mbf k\in \Gamma$ of $\mu^n$, that is
\begin{equation}
  \sum_{\mbf k \in \Gamma} d[\mbf k]\, \widehat{Df}(\bs\omega - 2\pi\mbf k) = 0, ~~\text{for all } \bs \omega \in \mathbb{R}^2.
  \label{eq:annihilation2}  
\end{equation}
\end{prop}
Likewise, by linearity, \eqref{eq:annihilation2} is valid for linear combinations $f = \sum_i g_i \cdot \chi_{\Omega_i}$, where $Dg_i = 0$ and $\mu = 0$ on $\cup_i \partial\Omega_i$. 

\section{Recovery from finite Fourier samples}


We now investigate necessary and sufficient conditions for the recovery of the filter coefficients describing the edge set from finitely many Fourier samples of the original signal $f$. For these results we restrict our attention to piecewise constant signals.

\subsection{{Necessary conditions}}
For a piecewise constant signal $f$, from the annihilation condition \eqref{eq:annihilation} we may form the linear system of equations:
\begin{equation}
\begin{cases}
  \sum_{\mbf k \in \Lambda} d[\mbf k] \widehat{f_x}\left(2\pi[\mbf l - \mbf k]\right) = 0,\\
  \sum_{\mbf k \in \Lambda} d[\mbf k] \widehat{f_y}\left(2\pi[\mbf l - \mbf k]\right) = 0,
\end{cases}
   ~~\forall ~\mbf l \in \Gamma.   
   \label{eq:annsys}
\end{equation}
where $\widehat{f_x}$ and $\widehat{f_y}$ may be computed from samples of $\widehat{f}$ by $\widehat{f_x}(\bs \omega) = -j\omega_x\cdot \widehat{f}(\bs\omega)$, and $\widehat{f_y}(\bs \omega) = -j\omega_y\cdot \widehat{f}(\bs\omega)$. Supposing the sampling grid $\Omega$ is a rectangular of dimensions $(K',L')$, and the minimal polynomial for $C$ has degree $(K,L)$, with coefficients $c[\mbf k]$ supported in $\Lambda$, then we may form at most $M = 2\cdot(K'-K +1)\cdot(L'-L+1)$ valid equations from \eqref{eq:annsys}. Therefore to solve for the at most $K\cdot L$ unknowns $c[\mbf k]$, $\mbf k \in \Lambda$, we require at least $M = K\cdot L$ equations. This gives the following necessary condition for recovery of $C$: 

\begin{prop}
\label{prop:necessity}
Let $f$ be piecewise constant such that the edge set $C$ has minimal polynomial $\mu$ of degree $(K,L)$. A necessary condition to recover the edge set $C$ from \eqref{eq:annsys}, is to collect samples of $\widehat{f}$ on a $(K',L')$ rectangular grid such that $$2\cdot(K'-K +1)\cdot(L'-L+1) \geq K\cdot L.$$
\end{prop}

\begin{figure}[!t]
\centering
\subfloat[Original signal]{\includegraphics[width=0.18\textwidth]{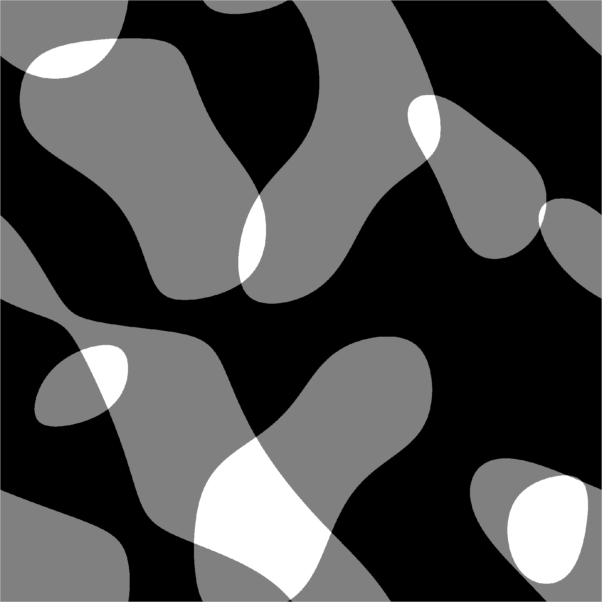}}
~~~
\subfloat[Recovered $\mu$]{\includegraphics[width=0.22\textwidth]{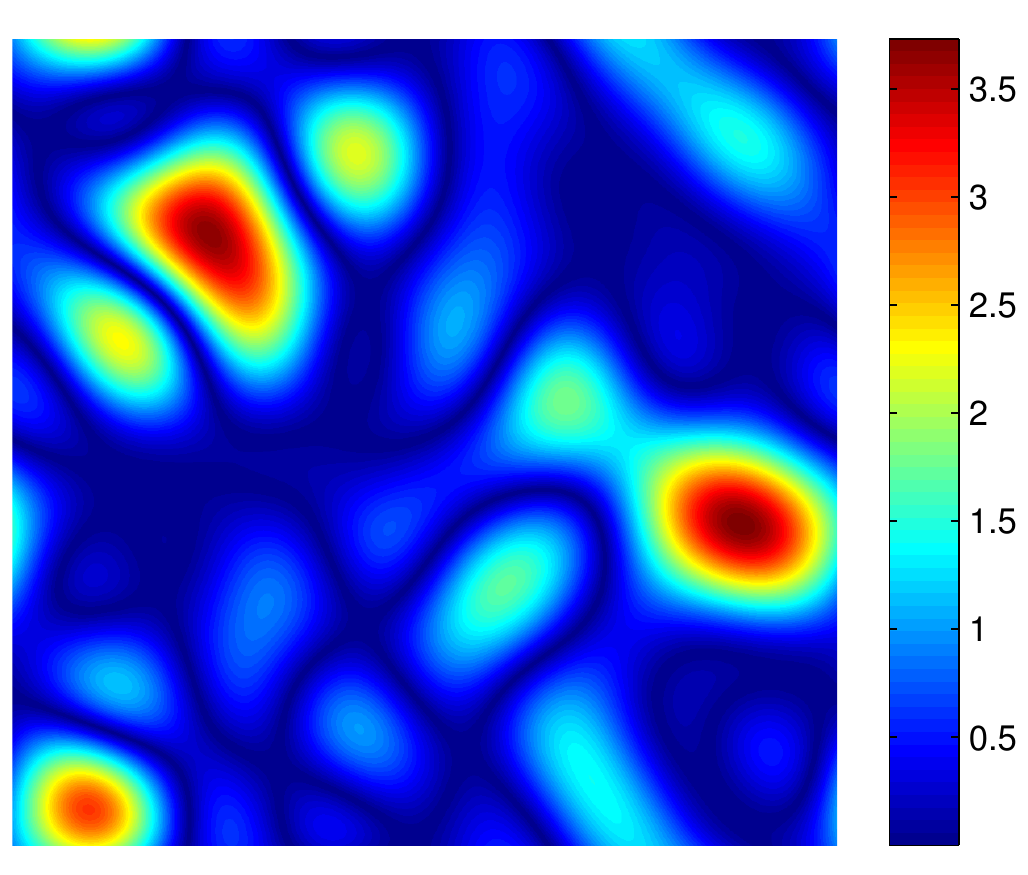}}\\
\subfloat[Edge set $\{\mu=0\}$]{\includegraphics[width=0.18\textwidth]{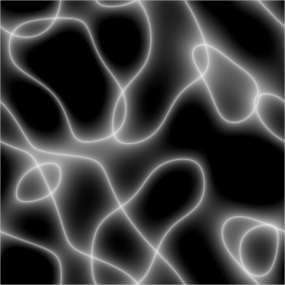}}
~~~
\subfloat[Singular values (log scale)]{\includegraphics[width=0.22\textwidth, trim=6mm 6mm 0mm 0mm,clip]{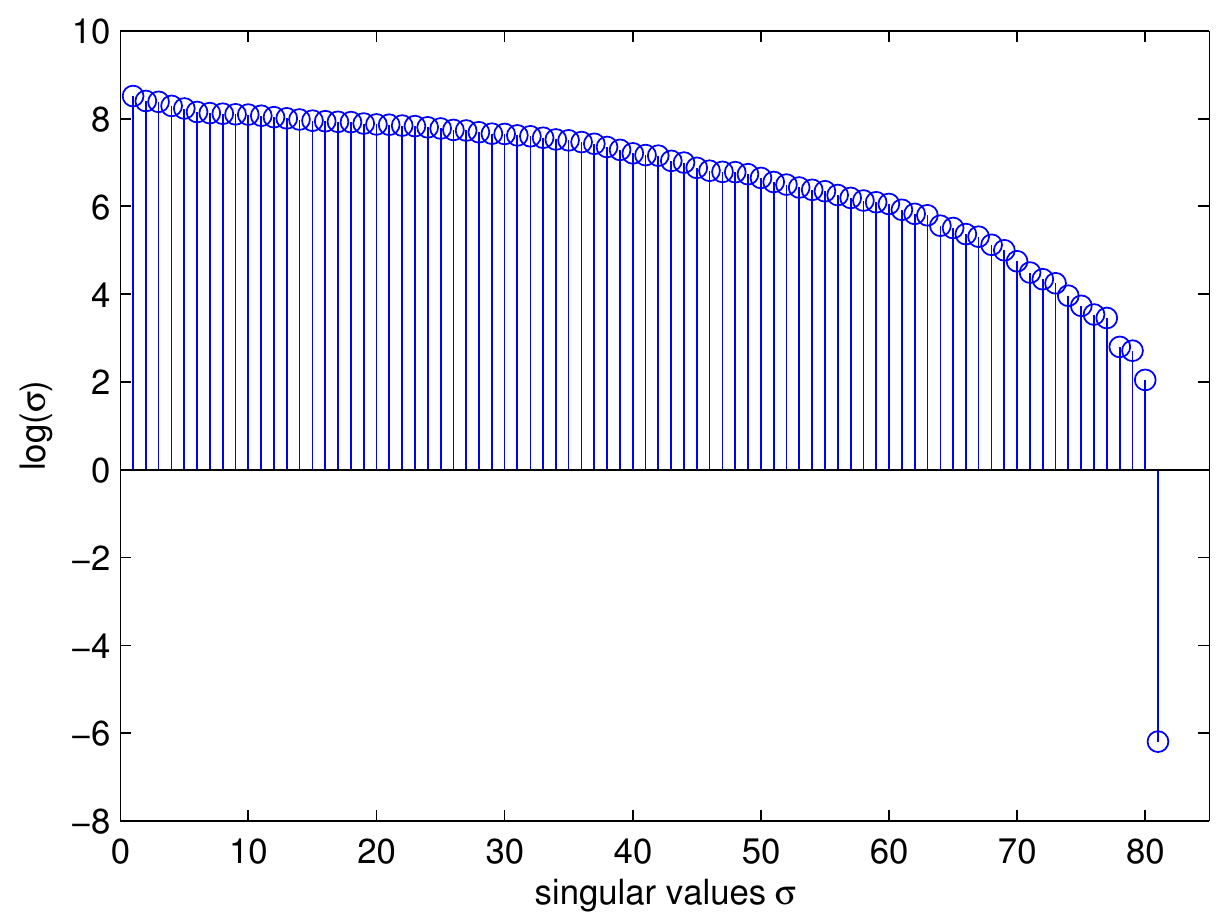}}
\caption{\small Exact recovery of edge set of a piecewise constant signal from the minimum necessary number of Fourier samples. The original piecewise constant signal is shown in (a), and was generated to have edge set $C:\{\mu =0\}$ where the degree of $\mu$ is known to be $(9,9)$. Here the minimal polynomial and edge set $C$ is shown to be recovered from the system \eqref{eq:annsys} having dimensions $81\times 81$ corresponding to the $15\times15$ necessary minimum number of Fourier samples predicted by Prop. \ref{prop:necessity}. The singular values of this system are plotted in (d), indicating the recovered $\mu$ is the only non-trivial solution.}
\label{fig:sim}
\end{figure}

To illustrate this bound, suppose the minimal polynomial has degree $(K,K)$, and we take Fourier samples from a square region. Then this requires at least $1.71K\times 1.71K$ Fourier samples to recover the edge set $C$. 
Our numerical experiments on simulated data (see Fig.\ \ref{fig:sim}) indicate the above necessary condition might also be \emph{sufficient} for unique recovery; that is, we hypothesize the minimal filter coefficients $c[\mbf k]$ are the only non-trivial solution to the system of equations \eqref{eq:annsys}.

\subsection{{Sufficient conditions}}We now focus on sufficient conditions for the recovery of the edge set. Here we will use $m\Lambda$ to denote a dilation of the set $\Lambda$ by a factor of $m$: if $\Lambda = \{\mbf (k,l): |k|\leq K, |l|\leq L\}$, then $m\Lambda = \{\mbf (k,l): |k|\leq m\, K, |l|\leq m\,L\}$.
\begin{thm}
\label{thm:unique1}
Let $f = \chi_\Omega$ be the characteristic function of a simply connected region $\Omega$ with boundary $\partial\Omega$ having minimal polynomial $\mu$ with coefficients $c[\mbf k],\mbf k \in \Lambda$.  Then the $c[\mbf k]$ can be uniquely recovered (up to scaling) as the only non-trivial solution to the equations
\begin{equation}
\begin{cases}
  \sum_{\mbf k \in \Lambda} c[\mbf k] \widehat{f_x}\left(2\pi[\mbf l - \mbf k]\right) = 0,\\
  \sum_{\mbf k \in \Lambda} c[\mbf k] \widehat{f_y}\left(2\pi[\mbf l - \mbf k]\right) = 0,
\end{cases}
   ~~\forall ~\mbf l \in 2 \Lambda.
  \label{eq:unique1}  
\end{equation}
\end{thm}
The proof entails showing any other trigonometric polynomial $\eta(r)$ having coefficients $d[\mbf k]$, $\mbf k \in \Lambda$ satisfying \eqref{thm:unique1} must vanish on $\partial\Omega$, from which it then follows that $\eta$ is a scalar multiple of the minimal polynomial $\mu$ by degree considerations. We also are able to show similar result holds for piecewise constant signals, provided the characteristic functions do not intersect:



\begin{thm}
\label{thm:unique2}
Let $f(\mbf r) = \sum_{i=1}^n a_i \chi_{\Omega_i}(\mbf r)$ be piecewise constant, where the boundaries $\partial\Omega_i$ are described by non-intersecting trigonometric curves $\{\mu_i = 0\}$, where $\mu_i$ is the minimal polynomial for $\partial\Omega_i$. Then, the coefficients $d[\mbf k]$,$k\in\Lambda$, of $\mu=\mu_1\cdots\mu_n$, and equivalently the edge set $C=\cup_{i=1}^n \partial \Omega_i$, can be uniquely recovered (up to scaling) as the only non-trivial solution of
\begin{equation}
\begin{cases}
  \sum_{\mbf k \in \Lambda} d[\mbf k] \widehat{f_x}\left(2\pi[\mbf l - \mbf k]\right) = 0,\\
  \sum_{\mbf k \in \Lambda} d[\mbf k] \widehat{f_y}\left(2\pi[\mbf l - \mbf k]\right) = 0,
\end{cases}
   ~~\forall ~\mbf l \in 2\Lambda.   
   \label{eq:unique2}
\end{equation}
\end{thm}

Note that to form the equations in \eqref{eq:unique1} and \eqref{eq:unique2} requires access to Fourier samples $\widehat{f}[\mbf k]$ for all $\mbf k\in 3 \Lambda$, which is greater than necessary number of samples given in Theorem \ref{prop:necessity}. We conjecture that the uniqueness results in Theorems \ref{thm:unique1} and \ref{thm:unique2} can in fact be sharpened to the necessary number of samples, and extended to piecewise constant signals where the boundaries of the regions intersect.

\section{Recovery Algorithms}
Up to now we have only considered the problem of recovering the edge set of a piecewise constant signal $f = \sum_{i=1}^n a_i \chi_{\Omega_i}$ from finite Fourier samples. In analogy with Prony's method, once the edge set is determined it is theoretically possible to recover the signal amplitudes $a_i$ by substuting $f$ back into \eqref{eq:unique2} and solving a full rank system. However, this is not feasible in practice since it requires factoring a high degree multivariate polynomial into its irreducible factors. Instead we pursue approaches that allow us to pose the recovery as the solution to a convex optimization problem.

\subsection{Curve-aware recovery}
Supposing we have access to an annihilating polynomial $\mu$ of the signal (equivalently, the edge set $C:\{\mu=0\}$), one approach is to pose the recovery as the weighted total variation minimization problem:
\begin{equation}
f = \arg\min_g \int |\mu(\mbf r)\,\nabla g(\mbf r)|\, d\mbf r \text{ subject to } \widehat{g}[\mbf k] = \widehat{f}[\mbf k], \forall \mbf k \in \Gamma
\label{eq:TVmin}
\end{equation}
Here, since $\mu = 0$ on the edge set $C$, the gradient of the image not penalized along the curve, which allows for the recovery of an image with sharp edges along $C$.

A version of this approach was investigated in an earlier work for the super-resolution recovery of MR signals from few Fourier samples \cite{isbi2015}. However, we found this scheme to have certain drawbacks, namely that it requires discretization onto a spacial grid, the optimization of many parameters, and is very sensitive to the estimate $\mu$. Hence we consider an alternate approach which does not rely on an explicit estimate of the annihilating polynomial $\mu$, but instead jointly recovers the image and the annihilating polynomial in a single stage algorithm. 

\subsection{Low-rank recovery}






The annhilation equations specified by \eqref{eq:annsys} can be represented in the matrix form as 
\begin{equation}
\underbrace{\left[\begin{array}{c}\mbf T_x[\widehat{f}]\\ \mbf T_y[\widehat{f}] \end{array}\right]}_{\mbf T[\widehat{f}]} \mbf d = \mbf 0
\label{matrixform}
\end{equation}
where $\mbf d$ is a vectorized version of the Fourier coefficients $d[\mbf k], \mbf k \in \Lambda$, and $\mbf T_x$ and $\mbf T_y$ are block Toeplitz matrices corresponding to the 2-D convolution of $d[\mbf k]$ with the discrete samples of $j\omega_x\,\widehat f(\bs\omega)$ and  $j\omega_y\,\widehat f(\bs\omega)$, respectively, for $\bs \omega = 2\pi \mbf l$, $\mbf l \in \Omega$. Specifically, if the coefficient support set $\Lambda$ has dimensions $K \times L$, each row of $\mbf T_x(\hat {f})$ is the vectorized version of an $K\times L$ patch of $j\omega_x\,\widehat f(\bs\omega)$, and likewise for $\mbf T_y(\hat {f})$. The number of rows is equal to the number of distinct patches, which correspond to the number of annihilation equations.

Note that for a given piecewise constant signal, a priori we do not know the degree of the minimal polynomial describing the edge set, which is needed to specify the size of $\mbf T$. However, if $d[\mbf k], \mbf k \in \Lambda'$ corresponding to the trigonometric polynomial $\mu$ is a solution to \eqref{matrixform} whose support set $\Lambda'$ is strictly smaller than the assumed support set $\Lambda$, then any multiple $\nu = \mu \cdot \gamma$ having coefficients $e[\mbf k] = (d \ast g)[\mbf k]$ supported within $\Lambda$, is also a solution. This implies that if we consider a larger filter size than required by the minimal polynomial, i.e. more columns in $\mbf T$ than the number of coefficients in $\mbf d$, the matrix $\mbf T$ will be low-rank.

The preceding discussion suggests we may pose the recovery of the signal as a structured low-rank matrix completion problem, entirely in the Fourier domain:
\begin{equation}
\widehat{f} = \arg\min_{\widehat{g}}~\text{rank}(\mbf T[\widehat{g}]) \text{ subject to } \widehat{g}[\mbf k] = \widehat{f}[\mbf k], \forall \mbf k \in \Gamma
\label{eq:rankmin}
\end{equation}
We note this approach is still ``off-the-grid'' in the sense that we may recover a discrete image at any desired resolution by extrapolating $\widehat{f}$ to this resolution in Fourier space, and applying an inverse DFT. To address the case of noisy measurements and model mismatch we propose solving the convex relaxation of \eqref{eq:rankmin}:
\begin{equation}
\widehat{f} = \arg\min_{\widehat{g}}~\|\mbf T[\widehat{g}]\|_* + \lambda\|P_{\Gamma}(\widehat{g} - \widehat{f})\|_2^2
\label{eq:nucnormmin}
\end{equation}
where $\|\cdot\|_*$ denotes the nuclear norm, i.e. the absolute sum of singular values, $\lambda$ is a tunable parameter, and $P_{\Gamma}$ is the projection onto the sampling set $\Gamma$. A standard approach to solving \eqref{eq:nucnormmin} is by an iterative singular value soft-thresholding algorithm, which requires an SVD of the estimate $T[\widehat{g}]$ at each step. Due to the size of $T[\widehat{g}]$, such an algorithm is computationally prohibitive in this case. Instead we use the SVD-free algorithm proposed in \cite{signoretto2013svd}, which involves introducing auxiliary variables $\mbf U \in \mathbb{C}^{M\times r}$ and $\mbf V \in \mathbb{C}^{N\times r}$ via the well-known relation $\|\mbf X\|_* = \min_{\mbf X = \mbf U \mbf V^H} \|\mbf U\|^2_F + \|\mbf V\|^2_F$, and enforcing the constraint $\mbf T[\widehat{g}] = \mbf U \mbf V^H$, with the ADMM algorithm. 

\begin{figure}[!t]
\centering
\subfloat[Fully sampled]{\includegraphics[width=0.16\textwidth]{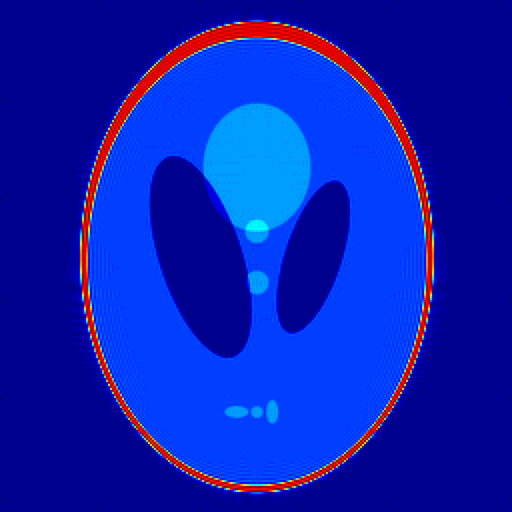}}
\subfloat[Zero-padded]{\includegraphics[width=0.16\textwidth]{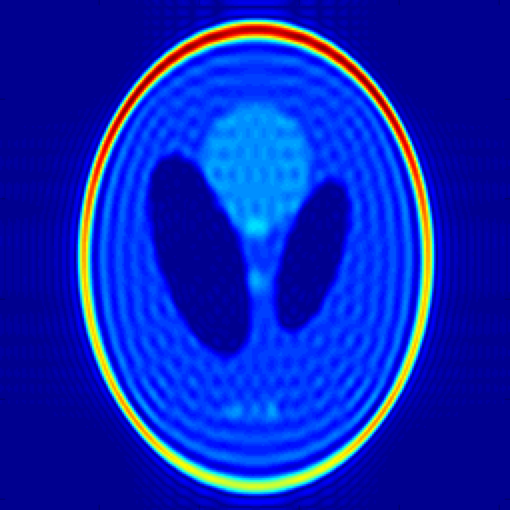}}
\subfloat[Low-rank]{\includegraphics[width=0.16\textwidth]{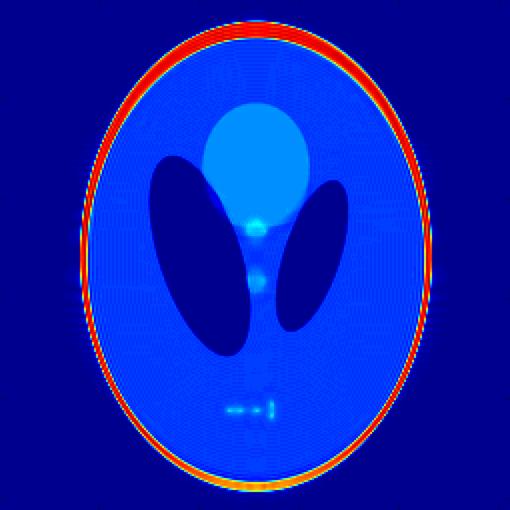}}
\vspace{-0.5em}
\subfloat{\includegraphics[width=0.16\textwidth]{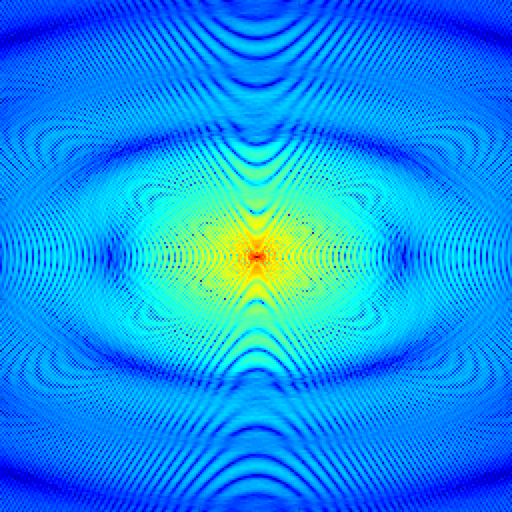}}
\subfloat{\includegraphics[width=0.16\textwidth]{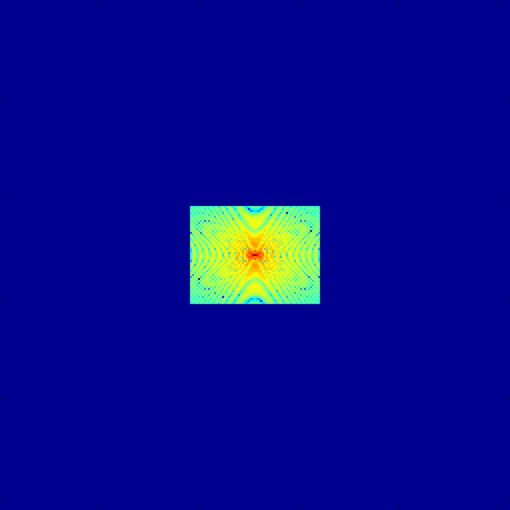}}
\subfloat{\includegraphics[width=0.16\textwidth]{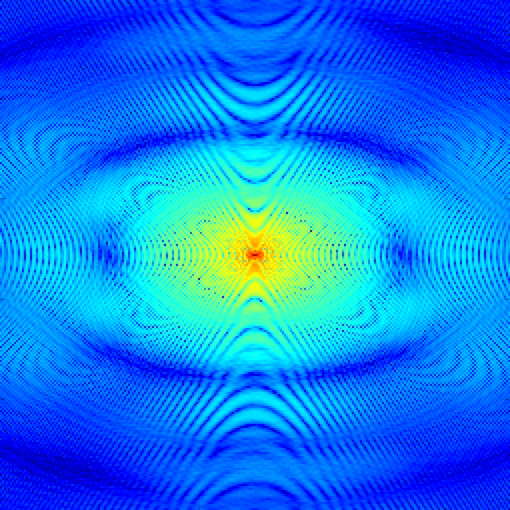}}
\caption{\small  Recovery of Shepp-Logan phantom on a 256x256 grid from 65x49=3185 Fourier samples ($\approx$20 fold undersampling). The top row shows the spatial domain images, while the bottom row shows the Fourier transforms of the images (log scale).}
\label{fig:lowrank_SL}
\end{figure}

In Fig.\ \ref{fig:lowrank_SL} we demonstrate the ability of the proposed algorithm to recover a piecewise constant signal from few of its uniform low-resolution Fourier samples. We experiment on simulated data obtained from analytical MRI phantoms derived in \cite{guerquin2012realistic}. 
We extrapolate from $65\times49=3185$ analytical Fourier samples of the Shepp-Logan phantom to a $256\times256$ grid ($\approx$20-fold undersampling), and recover the signal by performing a inverse DFT. Note that the ringing artifacts observed in the recovery are to be expected due to fact we are recovering exact Fourier coefficients of the signal, and could be removed with mild post-processing.


\section{Conclusion}
We propose an extension of the annihilating filter method to a wide class of 2-D piecewise smooth functions whose edges are supported on level set of a band-limited function. This enables us to recover an exact continous domain representation of the edge set from few low-frequency Fourier samples. In the case of piecewise constant signals, we derive conditions of the necessary and sufficient number of Fourier samples to ensure exact recovery of the edge set. Lastly, we prosed one-stage algorithm to recover piecewise constant signals by extrapolating the signal in Fourier domain. We demonstrate that we may accurately recover MRI phantoms from few of their low-resolution Fourier samples.





\bibliographystyle{IEEEtran}
\bibliography{IEEEabrv,root}
%

\end{document}